# Results of multi-agent system and ontology to manage ideas and represent knowledge in a challenge of creativity

Pedro Chávez Barrios
Escuela de Bachilleres, Universidad
Autónoma de Querétaro
Moises Solana s/n, 76070 Querétaro,
México
pedro.chavez@uaq.mx

Davy Monticolo
ERPI Laboratory, University of Lorraine
8 rue Bastien Lepage, 54000 Nancy,
France
davy.monticolo@univ-lorraine.fr

Sahbi Sidhom
LORIA (Kiwi) Laboratory, University of
Lorraine
F-54506 Vandœuvre lès Nancy, France
sahbi.sidhom@loria.fr

Abstract—**This article is about an intelligent system to support ideas management as a result of a multi-agent system used in a distributed system with heterogeneous information as ideas and knowledge, after the results about an ontology to describe the meaning of these ideas. The intelligent system assists participants of the creativity workshop to manage their ideas and consequently proposing an ontology dedicated to ideas. During the creative workshop many creative activities and collaborative creative methods are used by roles immersed in this creativity workshop event where they share knowledge. The collaboration of these roles is physically distant, their interactions might be synchrony or asynchrony, and the information of the ideas are heterogeneous, so we can say that the process is distributed. Those ideas are writing in natural language by participants which have a role and the ideas are heterogeneous since some of them are described by schema, text or scenario of use. This paper presents first, our MAS and second our Ontology design.**

## I. INTRODUCTION

The University of Lorraine organizes every year a creativity workshop called "48 hours generating ideas" (48H). We have observed that more than 1200 idea cards (IdC) were generated during the last 48H creativity workshop in 2017 [1]. **In order to manage these ideas a multi-agent system is studied and proposed since the multi-agent system has been proved to be efficient in a distributed process and to propose an ontology to represent knowledge.** The concept of multi-agent system appears at the end of last century. The multi-agent system has two forms of vision the interaction among agents and the interaction among humans, *the first, as an artificial intelligence (AI) concept attributed to Nils Nilsson "all AI is distributed-1980" and the second as artificial life (Alife) based in the complex adaptive behaviors of communities of humans* [2]. B*y relating an individual to a program, it is possible to simulate an artificial world populated of interacting processes* [3]. The individual is an agent that interacts according to his environment which is clearly defined with respect to the reality. These interactions among agents and their environments are an important aspect in the MAS. In the beginning of the century XXI, *an initial model tools used to create generic multi-agent platforms based on an organizational mode based in the core model agent-role-group* [4]*, and also multi-agent model involving some agents to hundreds focusing in break down a problem therefore the agent can solve a simple problem* [5]. At present, the multi-agent systems have been used to improve energy efficiency [6]. However, thinking in our intelligent system based in agents, the interactions among the actors in the creativity workshop 48H is complicated and we have to help them in their individual and collaborative activities inside this organization where *the organization is defined by a group of roles that interact among them* [7]. Several design methodologies of multi-agent system exist such as GAIA [8], [9] and DOCK [10] are examples of these design. The multi-agent systems have two principal methodologies, [11] the *methodologies oriented to agents and the methodologies oriented to organizations* that *base in organizational unit, service, environment and norm* [12]. Due to the uncountable times that agent is mention and the interaction of agent in a multi-agent system, we have to write some definitions about the concept of agent, it takes some primordial functions during the creation of our intelligent system. Since the last century and initial years of this century the concept of agent and its characteristics appears. *There are several definitions about agent, description of agent's requirements, uses of agent* [13] *and description about agent's evolution* [14]. In a software design an agent represents state components, which are structured aggregations *(sequences, sets, multisets, etc.) of elements such as events, actions, beliefs, plans and tasks* [15]. An actual definition, "Agent" is a system whose behavior is neither casual nor strictly causal, but teleonomic [16], "goal-oriented" toward a certain state of the world . an agent is specified as *an active communicating entity which plays roles inside groups*[4]. *An agent is a computer system* **situated** *in some* **environment**, *and that is capable of autonomous* **action** *in this environment in order to meet its design objectives where* **autonomous** *means to act without interventions of humans of other agents* [17].

**Having proposed the MAS next step is to define ontology**. With respect to the ontology, it is dedicated to ideas and specifically to assist participants in the idea generation during the creativity workshop. In addition, our ontology represents knowledge from this CWS like ideas, processes, activities, actors, roles, methods, idea cards and possible solutions; this ontology is used to annotate ideas and to facilitate the ideas management. As initial definition of ontology, the etymology of ontology comes from ancient

Greeks, but the concept of ontology appeared in the century XVII in the work "Ogdoas Scholastic" by Jacob Lorhard, he provided a useful key to the understanding of the Protestant Europe in a grammar text book [18]. Also, last century, the concept of ontology focus in the definition of objects, concepts, entities, relationships among them in a defined area [19], [20], and ontology works as database with information, properties, relationships about concepts that exist in the world or domain [21].

**The objective of this article is to present the results about our multi-agent and ontology proposals. Initially, we present the context, the problem and the methodology; the section two is about the state of art about MAS and Ontology approaches; the third section, our approaches. Finally, last section is dedicated to results and conclusions.**

## II. MULTI-AGENT SYSTEM (MAS) AND ONTOLOGY

### A. MAS and Ontology, state of the art

**Multi-agent system**

Inside the multi-agent system, the concept of agent is vital. The definition of agent in a general and complete AI idea: *an agent is anything that can be viewed as perceiving its environment through sensors and acting upon that that environment through effectors. A human agent has eyes, ears, and other organs for sensors, and hands, legs, mouth, and other body parts for effectors. A robotic agent substitutes cameras and infrared range finders for the sensors and various motors for the effectors* [22].

There are some varieties of agents but in a robotic sense: *Autonomous agents require be reactive to changes in the environment, it must be able to predict, incompatible goal management and adaptivity (prediction)* [23]. The agents, in our environment of creativity workshop 48H, is part of teams and can play one or more roles. Another definition, an agent is a physical or virtual entity that have several properties. These properties are the capacity to interact with its environment, the capacity of communication with other agents, the necessity to achieve an objective, the capacity to manage its resources, the capacity to perceive the environment, the capacity to represent partially or totally the environment and eventually the capacity of reproduction [24]**.** With a different Wooldridge's conceptions, an agent is an informatic system in a specific environment, with autonomous actions to achieve its objectives [25]. For a language, agent could be a mental state consisting in beliefs, desires and intentions [8]. A definition based in software, agent are coarse-grained computational system, each making use of computational resources, they are heterogeneous [8]. A Final definitions according to a multi agent system where agents interact among them to achieve a global objective, there exist two kind of agents, cognitive and reactive agents[24]. The artificial intelligence (AI) is an important discipline that defines agent in different ways. *AI borrows concepts (states, actions and rational agents) and techniques for autonomic computing. The definition of rational agent: is any entity that perceives and acts upon its environment, selecting actions that, on the basis of information from built-in knowledge, are expected to maximize de agent's objective* [26]. The most simple definition about agent, *A software agent has encoded bit strings as its percepts and actions* [22].

**Ontology**

In the last century, the concept of ontology focus in the definition of objects, concepts, entities, relationships among them in a defined area [19], [20], and ontology works as database with information, properties, relationships about concepts that exist in the world or domain [21].

Berners-Lee proposes to use the ontologies in the context of the Internet in order to bring a semantic dimension of the Web. He explains, "The Semantic Web will enable machines to comprehend semantic documents and data, semantic web uses collections of information called ontologies that is a component of the semantic web. Artificial-intelligence and Web researchers have co-opted the term for their own jargon, and for them an ontology is a document or file that formally defines the relationships among terms [27]". The semantic Web allow us to build ontologies by using a set of languages as RDF, RDFs and OWL to structure knowledge resources.

The creation of a domain ontology need to define in detail the concepts, the procedures, the activities and the relationships that belong specially to this domain or field trying to eliminate ambiguity and doubts due to the communication among web researches and machines (computers) using software applications, as Staab and Studer explain in (Staab and Studer, 2004).

Mathieu d'Aquin defines "Ontologies represent the essential technology that enables and facilitates interoperability at the semantic level, providing a formal conceptualization of the data which can be shared, reused, and aligned" [28]".

Elbassadi in [29] complete those definitions by explaining that the Ontologies provide a semantic representation of a common language to foster interoperability, declaratively, and intelligent services between tools and to support the innovation life cycle.

There are several existing libraries dedicated to ontology. An ontology library is a distributed data space where users and software agents can publish and access information from many different sources, the format RDF guarantees the interoperability making it possible for applications to reuse data and to link diverse data [30]. BioPortal is a library of biomedical ontologies developed by the National Center for Biomedical Ontologies, it provides essential domain knowledge to drive data integration, information retrieval, data annotation, natural-language processing and decision support [31].

### B. MAS Methodologies

There are some methodologies to design a multi-agent system (MAS). These methodologies involve mainly roles, agents, interactions among agents and the environment. First methodologies were developed at the end of the century based on interaction of roles [25] but in this century several methodologies appears such as ICTAM [32], DOCK methodology [33], MOBMAS [34], ADELFE [35], GAIA [36], [9].

**Wooldridge's GAIA Methodology**

The GAIA methodology is for agent-oriented analysis and design, in that it is applicable to a wide range of multi-agent systems, and comprehensive, in that it deals with both the macro-level (societal) and the micro-level (agent) aspects of systems. Gaia is founded on the view of a multi-agent system as a computational organization consisting of various interacting roles. Gaia has been specifically tailored to the analysis and design of agent-based systems [36].

**DOCK Methodology**

DOCK [37] helps to model a multi-agent system based in knowledge management. This methodology describes an intelligent knowledge system; it uses human organizations, roles, collaborations, skill, goals and knowledge. The methodology DOCK defines four elements: the organizational structure to identify agents (roles), the process model, the activity model and the role model. The model defines three stages: human organization, agent organization and interactions.

**MOBMAS Methodology**

This ontology-based methodology, used for the analysis and design of multi-agent systems [34]; MOBMAS is the first methodology that explicitly identifies and implements the various ways in which ontologies can be used in the MAS development process and integrated into the MAS model definitions. Conforming to the definition of a software engineering methodology [38], MOBMAS is comprised of a software engineering process that contains activities and associated steps to conduct the system development, techniques to assist the process steps and a definition of the models.

**GAIA methodology with abstractions**

This multi-agent system paradigm introduces a number of new abstractions and design/development issues. Gaia exploits the organizational abstractions to provide clear guidelines for the analysis and design of complex and open software systems [9].

**SABPO Methodology**

The SABPO methodology use an organizational metaphor, in which each agent plays a specific role to achieve the global goals of the organization, in addition, the methodology introduces a new interaction pattern. This approach puts the FIPA (the foundation for Intelligent Physical Agent) abstract architecture specification to the center of the methodology as a basic organizational structure and tries to create a concrete FIPA architecture that satisfies system requirements.

**ADELFE Methodology**

ADELFE is another methodology focus on the adaptive nature of the environment. It introduces the concepts of Non Cooperative Situations (NCS) that defines the behavior of local agents when they encounter an unpredictable situation in a dynamic environment. Rational Unified Process leads ADELFE that is devoted to software engineering adaptive MAS. ADELFE guarantees that the software is developed according to the adaptive multi-agent systems AMAS theory. The AMAS theory provides a solution to build complex systems for which classical algorithmic solutions cannot be applied, these systems are open and complex. All the interactions the system may have with its environment cannot be enumerated[35]. ADELFE is an agent-oriented methodology for designing Adaptive Multi-Agent Systems (AMAS), it is a French acronym for "Atelier de Développement de Logiciels à Fonctionalité Emergente" [39].

*C. Methodologies to design ontologies*

Ontology requires a well-defined process to represent the reality. Several methodologies already exist to build ontologies.

The construction of ontologies is very much an art rather than a science. This situation needs to be changed and will be changed only through an understanding of how to go about constructing ontologies. We will present in the next section some approaches to build ontologies.

**Enterprise Approach**

Uschold and King propose a methodology for ontology construction in [40]. This methodology according to Jones [41] bases on four steps:

- Identify the purpose determines the level of formality at which the ontology should be described.

- Identify the scope: a "Specification" is produced which fully outlines the range of information that the ontology must characterize. This may be done using motivating scenarios and informal competency questions, as in TOVE or by "brainstorming and trimming" i.e. produce a list of potentially relevant concepts and delete irrelevant entries and synonyms.

- Formalization: create the "Code", formal definitions and axioms of terms in the Specification.

- Formal evaluation: the criteria used may be general; this stage may cause a revision of the outputs of stages 2 identify the scope and 3 formalizations.

**Methontology**

The goal of this methodology, [42], is to clarify to readers interested in building ontologies, the activities they should perform and in which order, as well as the set of techniques to be used in each phase of the methodology. He thinks ontology is an art and tries to transform in an engineering. Here, the steps:

(1) Specification: identify the purpose of the ontology, including the intended users, scenarios of use, the degree of formality required, etc., and the scope of the ontology including the set of terms to be represented, their characteristics and the required granularity. The output of this phase is a natural-language ontology specification document.

(2) Knowledge acquisition: this occurs largely in parallel with stage (1). It is non-prescriptive as any type of knowledge source and any elicitation method can be used, although the roles of expert interviews and analyses of texts are specifically discussed.

(3) Conceptualization: domain terms are identified as concepts, instances, verbs relationships or properties and each are represented using an applicable informal representation.

(4) Integration: in order to obtain some uniformity across ontologies, definitions from other ontologies, e.g. Ontolingua standard units ontology, should be incorporated.

(5) Implementation: the ontology is formally represented in a language, such as Ontolingua.

6) Evaluation: much emphasis is placed on this stage in METHONTOLOGY. The techniques used are largely based on those used in the validation and verification of KBSs. A set of guidelines is given on how to look for incompletenesses, inconsistencies and redundancies.

(7) Documentation: collation of documents that result from other activities.

**KBSI IDEF5**

This method is devoted to assist in the creation, modification and maintenance of ontologies according to Jones in [41] and [43], the process of IDEF5:

(1) Organizing and scoping establishes the purpose, viewpoint, and context for the ontology development project. The purpose statement provides a set of "completion criteria" for the ontology, including objectives and requirements. The scope defines the boundaries of the ontology and specifies parts of the systems that must be included or excluded.

(2) Data collection: the raw data needed for ontology development is acquired using typical KA techniques, such as protocol analysis and expert interview.

(3) Data analysis: the ontology is extracted from the results of data collection. First, the objects of interest in the domain are listed, followed by identification of objects on the boundaries of the ontology. Next, internal systems within the boundary of the description can be identified.

(4) Initial ontology development: a preliminary ontology is developed, which contains proto-concepts i.e. initial descriptions of kinds, relationships and properties.

(5) Ontology refinement and validation: the proto-concepts are iteratively refined and tested. This is essentially a deductive validation procedure as ontology structures are "instantiated" with actual data, and the result of the instantiation is compared with the ontology structure.

**Methodology for ontology Ontolingua**

Ontolingua is mechanism for writing ontologies in a canonical format, such that they can be easily translated into a variety of representation and reasoning systems. This allows one to maintain the ontology in a single, machine-readable form while using it in systems with different syntax and reasoning capabilities. The syntax and semantics are based on the KIF knowledge interchange format [44]. Ontolingua extends KIF with standard primitives for defining classes, relationships, and organizing knowledge in object-centered hierarchies with inheritance.

The methodology used to design Ontolingua was:

- A well-defined, declarative semantics for all statements in the language.

- A mechanism that allows operational use of the ontologies within in a variety of implemented representation systems.

- A syntax that facilitates the modular definition of terms in an ontology, and the modular packaging of ontologies.

- A means for capturing conventions in knowledge representation and organization, such as class hierarchies and domain and range constraints on relationships, in a system independent, declarative form without sacrificing the efficient implementation of these conventions by various representation systems.

- An architecture and support library that makes it easy to write additional KIF translators.

**Methodology to design ontologies from organizational models**

The phases and activities applied to creativity workshops that represent the ontology process (cf. figure IV.37), it describes an ontology to model knowledge in creativity workshop, its description:

Phase 1: Definition

- Definition of domain, the Scope and Purpose.
- Definition of the questions-skills of the ontology (aptitudes).

Phase 2: Conceptualization

- Conceptualization and acquisition.
- The reuse of existing ontology concepts

Phase 3 Development

- The development of the ontology (programming, formalization).
- Population of the ontology

Phase 4: Validation/Evaluation

- Evaluation

III. MAS AND ONTOLOGY OUR APPROACH PROPOSALS

   A. Aplying MAS Methodology GAIA

The design processes of GAIA have three models, agent model, services model and acquaintance model that help us to understand the roles and interactions described before in the analysis phase. *Gaia is concerned with how a society of agents cooperate to realize the system-level goals, and what is required of each individual agent in order to do this. Actually how an agent realizes its services is beyond*

*the scope of Gaia, and will depend on the particular application domain* [36].

The models in GAIA are in two phases, first the analysis phase with the role model and interaction models and the second phase, the design phase with the agent model, service model and acquaintance model (cf. Figure 1).

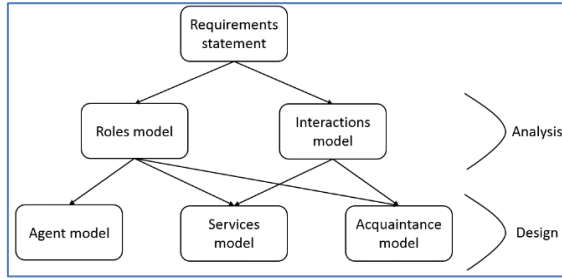

*Figure 1: Relations between the GAIA's models*[36] *p. 3.*

The objectives of a multi-agent system are to manage idea cards, to take decisions, and to enhance the creative techniques by means of the reactive and cognitive agents in the environment of 48H creativity workshop. The most abstract concept is the system. The organization is a collection of roles and interactions among them (Figure cf. 2). The analysis moves from abstract to concrete concepts.

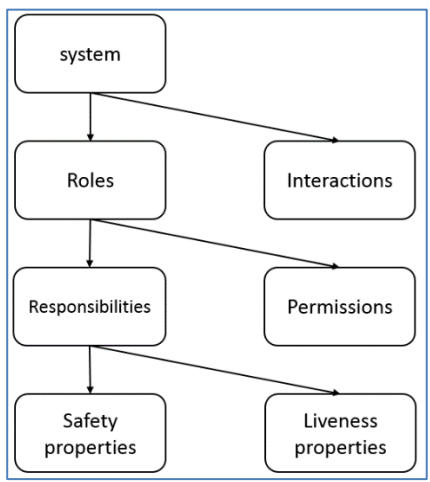

*Figure 2: Analysis of concepts* [36] *p. 4.*

### B. Applying the ontology Uschold ontologies

The proposed methodology, to build our ontology, must help to represents the evolution of the ideas, the individual ideas, after idea cards and finally possible solutions, all this evolution in an organizational model called 48H. This methodology follows a process based primordially on building the ontology. Initially, it expresses the meaning of the organization after it works on the design of ontology, finally the judgement and documentation.

The methodology chosed is the ontology of [45]. It has four phases:

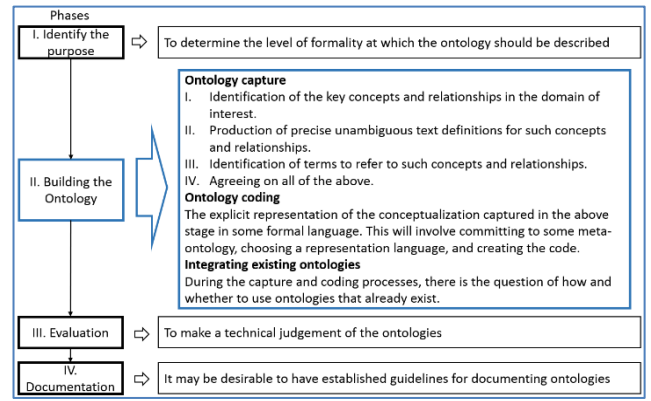

*Figure 3: Ontology Uschold phases.*

With this ontology, we can define easily the phases and identify the steps with the finality to do several iterations to correct our job. The Uschold ontology (cf. Figure 3) focus mainly in building the ontology that is something that we appreciate, for us the capture, coding and to integrate existing ontologies are vital steps without forget the iteration to improve.

### IV. RESULTS AND CONCLUSION

#### A. Multi-agent system results

**Analysis – Roles Model**

We will describe the roles of the agents also we have chosen to give the same role for the agents than for the participants of the creativity workshop i.e. "creative expert", "Industrial manager", "Organizer", "solver participant" and "technical expert". The objective of the agents is to assist the participants to achieve their activities and to manage their ideas during the creativity workshop process.

The schema of solver participant (cf. Table 1) details the production of ideas an idea cards. There are fifteen protocols (showed in the table 1 at the row of Protocols and activities) where solver participant acts. The permissions are producing an individual idea using individual activities, to produce at least two idea cards using a collaborative creative method, to have the same problem in WorkIdeaCards at the time of sharing with colleague team, to evaluate idea cards from the same problem except your idea card.

| Role Schema | Name |
|---|---|
| Description and objective | The role solver participant: To produce ideas in an individual way using activities. To produce idea cards by mean of collaborative creative methods. |
| Protocols and Activities | RequirementsInscription (Name, Last name, Institution), GiveRequirements (Name, Last name, Institution) Assignation (Assign_InstToWork, Assigned_ind, assigned_rol), Provides (part_team, problem) Offer_activity, SelectActivity, WorkIdeas Offer_method, SelectMethod, WorkIdeaCards, Improve, CompareIdeas, SendingIdeaCards, ReceivingPossibleSolutions, WatchingPossibleSolutions, AwardsEnd |
| Permissions | The actor must be register as a Solver Participant; To produce at least 1 individual idea using individual activities. |

|  | To produce 2 Idea Cards using a collaborative creative method. |
|  | To have the same problem in WorkIdeasCards at the time of sharing with colleague team. |
|  | To evaluate idea cards from the same problem except your idea cards and the idea cards from your team partner. |
|  | **Responsibilities** |
| Liveness | Solverparticipant = (RequirementsInscription.GiveRequirements)+ · (Assignation)+ · (Provides)+ · (Offer_activity.SelectActivity.WorkIdeas)+ · (Offer_method.SelectMethod.WorkIdeaCards.Improve)+ ·(CompareIdeas)+ · (SendingIdeaCards.ReceivingPossibleSolutions)+ · (WatchingPossibleSolutions.AwardsEnd)+ |
| Safety | Idea > 0<br>Idea Card = 2 by team. |

*Table 1: Role Solver Participant.*

**Analysis – Interaction Model**

This second model describes the communication's protocols for each agent. The agent's protocol has some elements (cf. Figure 4) that help us to improve the explanation about the protocol's description (Colleman et al. 1996) in the interaction of the agents.

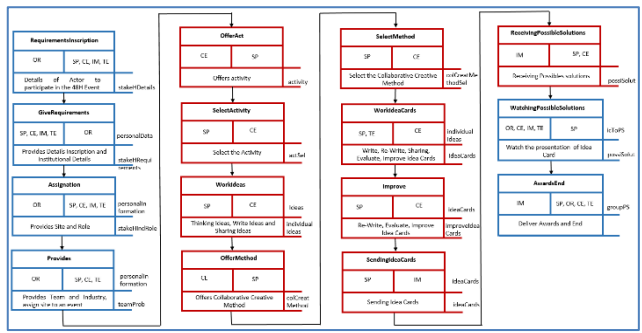

*Figure 4: Definition of protocols associated with Role Solver Participant.*

**Design – Agent Model**

In the agent model (cf. Figure 5), we identify seven agents and their instances that will make up the system. During the creativity workshop, it identifies easily the five roles according to the agent model proposed. The role creative expert and Technical Expert will form an agent called Creative Technological Expert Agent (CTEAgent-CTEA), the number of CTEA agents are 1 or more. The rest of the roles have their agents, making note that the agent Organizer (ORAgent-ORA) has one or more instances. However, we add three agent, semantic model knowledge agent (SMKA), width semantic distance agent (WSDA) and the comparative similarity agent (CSA). These agents help us to order ideas according to semantic distance, width-density and comparative similarity of Idea Cards. The definitions of the agents are:

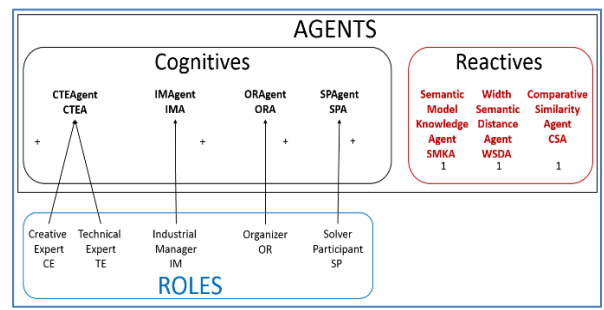

*Figure 5: The Agent Model.*

**Design – Service Model**

The services model identifies the main services that are required to realize the agent's role. These services (cf. Table 2) are functions that the agents have to execute according to the protocols described before.

| Service | Inputs | Outputs | Pre-condition | Post-condition |
|---|---|---|---|---|
| Obtain information of actors and assignation of roles | Actor details Name, last name, institution, sex, date of birth | Actor requirements | Event=1 | Institution>=1 Industry>=1 Role>=2 Team>=2 Problem>=1 |
| Selection and application of activity for ideas | Group, Creative Technical Expert, Activities | Ideas | Ideas per participant at least in mind | Idea>0 |
| Selection and Application of Methods for idea Cards | Thousands of Ideas, many methods | Idea Cards | 2 Idea Cards per group =2 | Idea Cards >2 |
| Evaluation by partners and improving idea card as a goal | Two ideas per group | Idea Cards | 2 Idea Cards per group | Idea Cards > 2 |
| Classification of Idea Cards | n Idea Cards | n Idea Cards | At least 2 Idea Cards | Idea Cards >n |
| Sending Possible Solutions | Idea Cards | Possible Solutions | 2 possible solution per group | Possible solution >=2 |

*Table 2: Service Model.*

**Design – Acquaintance Model**

The acquaintance model, (cf. Figure 6), documents the lines of communication among the different agents. The agents Creative Expert CTEA, Organizational ORA, Solver Participant SPA and Industrial Manager have communication among them during the entire creativity workshop, the agents CTEA and SPA has relation with the agents Semantic Distance SDA, Width Density WDA and Comparative Similarity CSA. The agents SDA, WDA and CSA take action with the purpose of classify the ideas at the end of the creation, evaluation (among partner group and the rest of the groups) and improving.

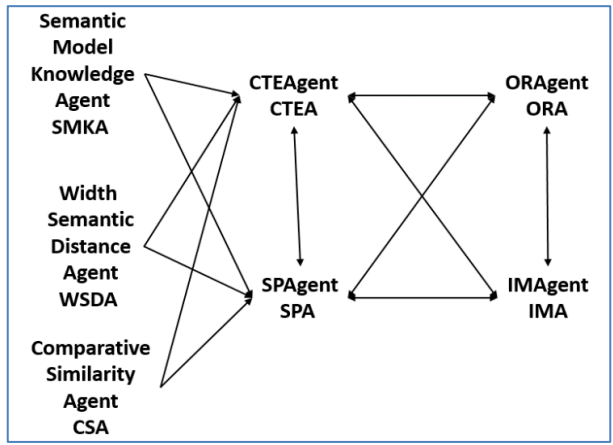

*Figure 6: The Acquaintance Model.*

### B. Ontology results

The Domain is an ontology dedicated to the creqtivity and we called it "The Collaborative Creative Ideas Ontology" CCIDEAS.

The scope describes, specifies and represents all the concepts relate to the creativity workshop 48 hours challenge. These concepts are identified inside of an organizational model.

The purposes of this ontology are to represent knowledge and ideas and to understand the creativity workshop. The ontology will be used to compare ideas in the intelligent system.

Creation of concepts
These concepts and relationships determine the ontology.
The definition of every concept that are part during this creativity workshop (*cf.* Table IV.16); this definition will be used to create triplets among concepts.

| Name of concept | Definition |
|---|---|
| 1 Activity | The action(s) that actor follows to produce ideas (I) in the phase of divergence. The activities of divergence used to produce individual ideas. Type: String |
| 2 Actor | The person that will participate in the event and can take a role to solve problems. The concept will indicate the role or several roles to assume during a CWS. Type: String |
| 3 Collaborative creative method (CCM) | Set of instructions applied by solver participants to generate idea cards (IC). The methods of convergence are used to produce Idea cards. Type: String. |
| 4 Event | The name of the CWS and its edition. Type: String |
| 5 Idea | The individual ideas (I) are produced in the phase of divergence; they are created using combinational, exploratory and transformational techniques with individual activities. The actor captures the initial individual idea. Type: String |
| 6 IdeaDesc | Idea's description. Type: String |
| 7 IdeaCard | The result of the use of ideas (I) and collaborative creative methods (CCM). An actor with the role of solver participant and from a team creates idea cards using CCM. |
| 8 ICDesc | Idea Card's description. Type: String. |
| 9 ICTitle | Idea Card's title. Type: String. |
| 10 ICScenery | Idea Card's scenery. Type: String. |
| 11 ICPrioCli | Idea Card's priority client. Type: String. |
| 12 ICAdvant | Idea Card's advantage. Type: String. |
| 13 ICRisk | Idea Card's risk. Type: String. |
| 14 Industry | The name of the industry, this concept has the problem. The industrial manager proposes the industry that contains the problem. Type: String |
| 15 Problem | The reasons why the organization creates the creativity workshop CWS events every year. Industries like Assystem, Bostik, CEA Tech, Decathlon, GRDF, ICM, MSA Safety, Muller, Normande Aerospace, Pierre Fabre, and Scarabée Biocop participate in those events. The problem is assigned to a team by mean of the industry. Type: String |
| 16 Role | The type of character that the actor takes. (Organizer, solver participant, creative expert, technical expert, Industrial manager). Type: String |
| 17 Site | The place where actors will work (ASU BAHRAIN, CESI NANTERRE, ENSGSI NANCY, etcetera). The site given to an actor. Type: String |
| 18 Team | The set of actors with the same role (solver participant), event, problem, site, name of team and colour. (Str_Ass_1, Lyo_Assy_1, Uca1, Str_Ass_2, etcetera). An actor takes part of a team. Type: String |
| 19 Vocabulary | The vocabulary is formed by ADJECTIVE, ADVERB, NOUN, VERB, ARTICLE, PRONOUN, PREPOSITION, CONJUNCTION AND INTERJECTION; some fields of the idea card's concept use these concepts such as title, description, priority client, name, scenery, advantages and risk. The vocabulary is part of idea's description. Type: String |
| 20 Organizer | The actor who takes the role in activities of assignation of roles, industries, team and event. Initially, this role create the event and he is asking for information to the actors with the purpose to do the inscription. Type: String |

| 21 Solver Participant | This concept is part of the roles and part of the team. The site is assigned to the solver participant; other relationships are: Select activity, Create ideas, Send possible solutions. Type: String |
|---|---|
| 22 Creative Expert | The creative expert offers activities and collaborative creative methods to the team of solver participants. Type: String |
| 23 Technical Expert | Technical expert helps teams to improve idea cards. Type: String |
| 24 Industrial Manager | The industrial manager proposes an industry to the creativity workshop. Type: String |
| 25 Possible Solutions | The concept possible solutions are the idea cards with best score according to the semantic skills like width, semantic distance and similarity. |

*Table 3: Definition of concepts.*

Creation of relationship proposed

The creation of relationships uses the format subject-verb-object with the purpose of create sets of three elements; *the term of relationships the concepts (Bachimont 2000) where the relationship represents the verb* (*cf.* Table 4 in the Annexe). Finally, we present the global ontology figure 7 in the annexe too.

V. CONCLUSION

*A. Multi-agent system and ontology*

**Contributions**

Our main contribution with the MAS and ontology is to assist participants of the creativity workshop to manage their ideas and knowledge and to propose an intelligent system based in multi-agent and an to develop an annotation system (ontology).

The agents inside MAS manage ideas but probably in the future other agents could be focused to assist participants in the selection of a better activity or collaborative creative method.

**References:**

[1] "Ecole d'été RRI 2017, les 28-29 août 2017: ' L'innovation agile : Quels défis pour les individus, les organisations et les territoires ?' - soutenue par IDEFI - InnovENT-E Nancy, Grande Région Est, France. [Online]. Available: https://rni2017.event.univ-l."

[2] D. Weyns, V. D. H. Parurak, and F. Michel, *Environments for Multi-Agent Systems, Lecture Notes in Artificial Intelligence*. 2004.

[3] A. Drogoul, J. Ferber, and C. Cambier, "Multi-agent Simulation as a Tool for Analysing Emergent Processes in Societies," pp. 49–63, 1992.

[4] O. Gutknecht and J. Ferber, "MadKit: A generic multi-agent platform," *Proc. fourth Int. Conf. Auton. agents - AGENTS '00*, pp. 78–79, 2000.

[5] O. Simonin and J. Ferber, "Un modèle multi-agent de résolutio collective de problèmes situés multi-échelles," pp. 1–13, 2003.

[6] W. Zhang, W. Liu, X. Wang, L. Liu, and F. Ferrese, "Distributed multiple agent system based online optimal reactive power control for smart grids," *IEEE Trans. Smart Grid*, vol. 5, no. 5, pp. 2421–2431, 2014.

[7] J. Ferber, "Coopération réactive et émergence," pp. 19–52, 1994.

[8] M. Wooldridge, M. Fisher, M. Huget, and S. Parsons, "Model Checking Multi-Agent Systems with MABLE ∗," pp. 952–959, 2002.

[9] F. Zambonelli, N. R. Jennings, and M. Wooldridge, "Developing Multiagent Systems: The Gaia Methodology," *ACM Trans. Softw. Eng. Methodol.*, vol. 12, no. 3, pp. 317–370, 2003.

[10] J. Girodon, D. Monticolo, E. Bonjour, and M. Perrier, "An organizational approach to designing an intelligent knowledge-based system: Application to the decision-making process in design projects," *Adv. Eng. Informatics*, vol. 29, no. 3, pp. 696–713, 2015.

[11] S. Esparcia, E. Argente, and V. Botti, "An agent-oriented software engineering methodology to develop adaptive virtual organizations," *IJCAI Int. Jt. Conf. Artif. Intell.*, no. i, pp. 2796–2797, 2011.

[12] E. Argente, V. Julian, and V. Botti, "MAS Modeling Based on Organizations," *Agent-Oriented Softw. Eng. IX 9th Int. Work.*, pp. 16–30, 2009.

[13] W. Shen and D. H. Norrie, "Agent-Based Systems for Intelligent Manufacturing: A State-of-the-Art Survey," *Knowl. Inf. Syst.*, vol. 1, no. 2, pp. 129–156, 1999.

[14] L. Vanhée, J. Ferber, and F. Dignum, "Agent-Based Evolving Societies ( Extended Abstract ) Categories and Subject Descriptors," pp. 1241–1242, 2013.

[15] D. Kinny, "A Visual Programming Language for Plan Execution Systems," *Proc. First Int. Jt. Conf. Auton. Agents Multi-Agent Syst. (AAMAS-2002, Featur. 6th AGENTS, 5th ICMAS 9th ATAL), 15--19 July, Bol. Italy*, pp. 721–728, 2002.

[16] C. Castelfranchi, "Guarantees for autonomy in cognitive agent architecture," pp. 56–70, 2012.

[17] N. R. Jennings and M. Wooldridge, *Applications of Inteligent Agents. Agent Technology: Formations, Applications and Markets.* 1998.

[18] P. Øhrstrøm, H. Shârphe, and S. Uckelman, "Jacob Lorhard's Ontology: A 17th Century Hypertext on the Reality and Temporality of the World of Intelligibles," in *Conceptual Structures: Knowledge, Visualization and Reasoning.Lecture Notes in Artificial Intelligence.*, 2008, pp. 74–87.

[19] T. R. Gruber, "Toward Principles for the Design of Ontologies," *International journal of human-computer studies*, vol. 43, no. 5. pp. 907–928, 1995.

[20] M. R. Genesereth and N. J. Nilsson, "Logical Foundations of Artificial Intelligence: Nomotonic reasoning," *Logical Foundations of Artificial Intelligence*. 1987.

[21] K. Mahesh, "Ontology development for machine translation: Ideology and methodology," *Comput. Res. Lab. New Mex. State Univ. MCCS-96-292*, p. 87, 1996.

[22] S. Russell and P. Norvig, *Artificial Intelligence: A Modern Approach.*, 2nd. Ed. 2003.

[23] V. Decugis and J. Ferber, "Action selection in an autonomous agent with a hierarchical distributed reactive planning architecture," *AGENTS '98 Proc. Second Int. Conf. Auton. agents*, pp. 354–361, 1998.

[24] J. Ferber, *Les systèmes Multi-Agents vers une Intelligence Collective*, InterEditi. 1995.

[25] M. Wooldridge, N. R. Jennings, and D. Kinny, "A Methodology for Agent-Oriented Analysis and Design: A Conceptual Framework," *Proc. Third Int. Conf. Auton. Agents*, pp. 69–76, 1999.

[26] J. O. Kephart and W. E. Walsh, "An artificial intelligence perspective on autonomic computing policies," *Proc. - Fifth IEEE Int. Work. Policies Distrib. Syst. Networks, POLICY 2004*, pp. 3–12, 2004.

[27] T. Berners-Lee and J. Hendler, "The Semantic Web," *Sci. Am.*, vol. 21, 2002.

[28] M. D'Aquin and N. F. Noy, "Where to publish and find ontologies? A survey of ontology libraries," *J. Web Semant.*, vol. 11, pp. 96–111, 2012.

[29] L. Elbassiti and R. Ajhoun, "Semantic Representation of Innovation, Generic Ontology for Idea Management," *J. Adv. Manag. Sci.*, vol. 2, no. 1, pp. 128–134, 2014.

[30] M. D'Aquin and N. F. Noy, "Where to publish and find ontologies? A survey of ontology libraries," *J. Web Semant.*, vol. 11, pp. 96–111, 2012.

[31] N. F. Noy *et al.*, "BioPortal : ontologies and integrated data resources at the click of a mouse," vol. 37, no. May, pp. 170–173, 2009.

[32] S. Elsawah, J. H. A. Guillaume, T. Filatova, J. Rook, and A. J. Jakeman, "A methodology for eliciting, representing, and analysing stakeholder knowledge for decision making on complex socio-ecological systems: From cognitive maps to agent-based models," *J. Environ. Manage.*, vol. 151, pp. 500–516, 2015.

[33] J. Girodon, D. Monticolo, and E. Bonjour, "How To Design a Multi Agent System Dedicated to Knowledge Management ; the DOCK Approach," pp. 113–121, 2015.

[34] Q. N. N. Tran and G. Low, "MOBMAS: A methodology for ontology-based multi-agent systems development," *Inf. Softw. Technol.*, vol. 50, no. 7–8, pp. 697–722, 2008.

[35] G. Picard and M. Gleizes, "THE ADELFE METHODOLOGY," in *Methodologies and Software Engineering for Agent Systems - The Agent-Oriented Software Engineering Handbook*, 2006, pp. 1–2.

[36] M. Wooldridge, N. R. Jennings, and D. C. N.-W.-2000-01 Kinny, "The Gaia Methodology for Agent-Oriented Analysis and Design," *J. Auton. Agents Multi-Agent Syst.*, vol. 3, no. 3, pp. 285–312, 2000.

[37] A. Gabriel, "Gestion des connaissances lors d'un processus collaboratif de créativité. Université de Lorraine, ERPI Lab. Décembre 2016.," 2016.

[38] B. Henderson-Sellers, A. Simmons, and H. Younessi, *The OPEN Toolbox of Techniques*. UK, 1998.

[39] S. Rougemaille, J. P. Arcangeli, M. P. Gleizes, and F. Migeon, "ADELFE design, AMAS-ML in action: A case study," *Lect. Notes Comput. Sci. (including Subser. Lect. Notes Artif. Intell. Lect. Notes Bioinformatics)*, vol. 5485 LNAI, pp. 105–120, 2009.

[40] M. Uschold and M. King, "Towards a Methodology for Building Ontologies," *Methodology*, vol. 80, no. July, pp. 275–280, 1995.

[41] D. Jones, T. Bench-Capon, and P. Visser, "Methodologies for Ontology Development," no. November 2012, 1998.

[42] M. Fernandez, A. Gomez-Pérez, and N. Juristo, "METHONTOLOGY : From Ontological Art Towards Ontological Engineering," p. 40, 1997.

[43] B. Peraketh *et al.*, "Ontology Capture Method (IDEF5)," 1994.

[44] M. R. Genesereth, R. E. Fikes, and T. Gruber, "Knowledge Interchange Format," *Interchange*, no. January, 1992.

[45] M. Uschold and M. King, "Towards a Methodology for Building Ontologies," *Methodology*, vol. 80, no. July, pp. 275–280, 1995.

**Annexe**

| Relation name proposed | Domains (Concepts) | Range (Concepts) | Triplet and/or Definition |
|---|---|---|---|
| 1 Select | Solver Participant | Activity. Examples: Brainstorming, write storming, Bend it and Shape it, Brain borrow, Copy cat, … | Solver Participant selects Activity. The property indicates that the Solver Participant select an Activity to create individual ideas during the phase of divergence |
| 2 Offers | Creative Expert | Activity. Examples: Brainstorming, write storming, Bend it and Shape it, Brain borrow, Copy cat, … | Creative Expert offers Activity. The property indicates that Creative offers an activity. |
| 3 Plays | Actor | Role. Examples: Creative Expert, Technical Expert, Industrial Manager, Solver Participant and Organizer | Actor plays a Role. |
| 4 Assign | Organizer | Role. Examples: Creative Expert, Technical Expert, Industrial Manager and Solver Participant. | Organizer assign Role. The property indicates that Organizer assigns all the roles in the creativity workshop. |
| 5 Propose | Industrial Manager | Industry. Examples: Decathlon, ICM, Bostik, etc. | Industrial Manager proposes an industry. The property indicates that Industrial Manager proposes an industry. |
| 6 Create | Organizer | Event. Examples: 48h InnovENT-Edition 2016, Operation 2015 InnovENT-E 48 hours to bring ideas to life. | Organizer creates an Event. |
| 7 Assign | Organizer | Site. Examples: INSA LYON, ENSGSI, UCA MARRAKECH, etc. | Organizer assigns Site. |
| 8 Assign | Organizer | Industry; Examples: Examples: Decathlon, ICM, Bostik, etc. | Organizer assigns Industry. |
| 9 Assign | Organizer | Team. Examples: Nan_Dec_1, Nan_Dec2, Str_Ass_2, etc. | Organizer assigns Teams. |
| 10 Requires | Organizer | Actor. Examples: Any institutional, educative or industrial person interested in creativity and solving problems. | Organizer requires Actor. |
| 11 Help | Technical Expert | Team. Examples: Nan_Dec_1, Nan_Dec2, Str_Ass_2, etc. | Technical Expert helps Team. The property indicates that Technical experts helps teams. |
| 12 IsAssignedTo | Industry | Team. Examples: Nan_Dec_1, Nan_Dec2, Str_Ass_2, etc. | Industry is assigned to Team. |
| 13 Receive | Industrial Manager | Possible Solutions | Industrial Manager receives Possible Solutions. |

|  |  |  |  | The property indicates that Industrial Manager receives the possible Solutions. |
|---|---|---|---|---|
| 14 | IsPartOf | Actor | Team. Examples: Nan_Dec_1, Nan_Dec2, Str_Ass_2, etc. | Actor is part of Team. |
| 15 | IsAssignedTo | Site | Event. Examples: 48h InnovENT-Edition 2016, Operation 2015 InnovENT-E 48 hours to bring ideas to life. | Site is assigned to an Event. |
| 16 | Send | Solver Participant | Possible Solutions | Solver Participant sends Possible Solutions. The property indicates that Solver Participant sends the possible Solutions. |
| 17 | Present | Team | Possible Solutions | Team presents Possible Solutions. The property indicates that Team presents the possible Solutions. |
| 18 | IsAssignedTo | Site | Role. Range: Technical Expert, Solver Participant and Creative Expert. | Site is assigned to Role. |
| 19 | IsAssignedTo | Team | Role. Range: Technical Expert, Solver Participant and Creative Expert. | Team is Assigned to Role. |
| 20 | Create | Team | Idea Card | Team creates Idea Card. The property indicates that Team creates the Idea Cards. |
| 21 | Improve | Team | Idea Card | Team improves Idea Card. The property indicates that Team improves the Idea Cards. |
| 22 | Select | Team | CCM. Examples: Six hats of thinking, The shirt off your back, Puzzle pieces, Organizational brainstorms, Best off, Rice storm, … | Team select CCM. The property indicates that Team selects the Collaborative Creative Method. |
| 23 | Use | CCM | Idea. | CCM uses Ideas. |
| 24 | Form | Idea | Idea Cards | Ideas form Idea Card. |
| 25 | Use | Idea Card | CCM. Examples: Six hats of thinking, The shirt off your back, Puzzle pieces, Organizational brainstorms, Best off, Rice storm, … | Idea Card uses CCM. |
| 26 | Offer | Creative Expert | CCM. Examples: Six hats of thinking, The shirt off your back, Puzzle pieces, Organizational brainstorms, Best off, Rice storm, … | Creative Expert offers CCM. |
| 27 | IsPartOf | IdeaDesc | Idea | IdeaDesc is part of Idea. The property indicates that Idea Description (IdeaDesc) is part of the Idea. |
| 28 | Create | Solver Participant | Idea | Solver Participant creates Idea. The property indicate that Solver Participant creates ideas. |
| 29 | IsPartOf1 | ICDesc | Idea Card | ICDesc is part1 of idea card. The property indicates that the field Idea Card Description (ICDesc) is part of the Idea Card. |
| 30 | IsPartOf2 | ICTitle | Idea Card | ICTitle is part2 of idea card. The property indicates that the field Idea Card Title (ICTitle) is part of the Idea Card. |
| 31 | IsPartOf3 | ICScenery | Idea Card | ICScenery is part3 of idea card. |

| | | | |
|---|---|---|---|
| | | | The property indicates that the field Idea Card Scenery (ICScenery) is part of the Idea Card. |
| 32 IsPartOf4 | ICPrioCli | Idea Card | ICPrioCli is part4 of idea card. The property indicates that the field Idea Card Priority Clients (ICPrioCli) is part of the Idea Card. |
| 33 IsPartOf5 | ICAdvant | Idea Card | ICAdvant is part5 of idea card. The property indicates that the field Idea Card Advantage (ICAdvant) is part of the Idea Card. |
| 34 IsPartOf6 | ICRisk | Idea Card | ICRisk is part6 of idea card. The property indicates that the field Idea Card Risk (ICRisk) is part of the Idea Card. |

*Table 4: Definition of relationships*

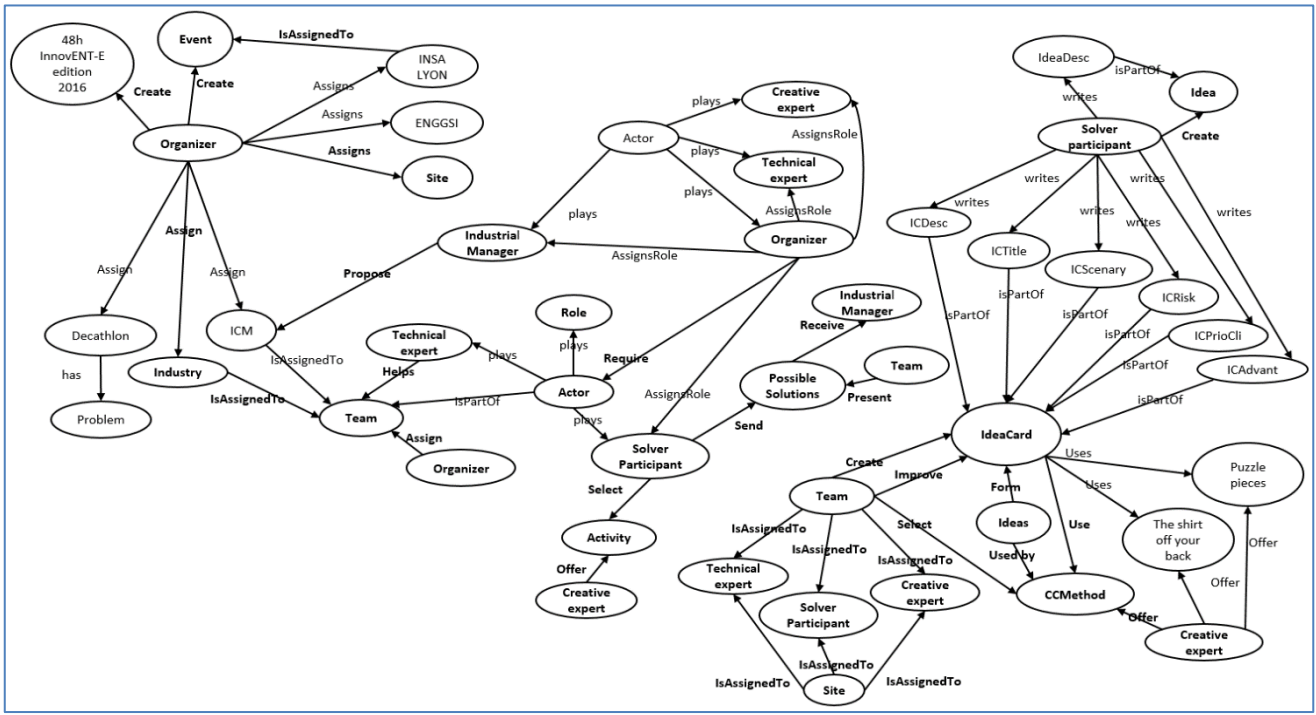

*Figure 7: Global Ontology.*